# PixelBoost: Leveraging Brownian Motion for Realistic-Image Super-Resolution

Aradhana Mishra and Bumshik Lee *Member, IEEE*

*Abstract*— Diffusion-model-based image super-resolution techniques often face a trade-off between realistic image generation and computational efficiency. This issue is exacerbated when inference times by decreasing sampling steps, resulting in less realistic and hazy images. To overcome this challenge, we introduce a novel diffusion model named PixelBoost that underscores the significance of embracing the stochastic nature of Brownian motion in advancing image super-resolution, resulting in a high degree of realism, particularly focusing on texture and edge definitions. By integrating controlled stochasticity into the training regimen, our proposed model avoids convergence to local optima, effectively capturing and reproducing the inherent uncertainty of image textures and patterns. Our proposed model demonstrates superior objective results in terms of learned perceptual image patch similarity (LPIPS), lightness order error (LOE), peak signal-to-noise ratio (PSNR), structural similarity index measure (SSIM), as well as visual quality. To determine the edge enhancement, we evaluated the gradient magnitude and pixel value, and our proposed model exhibited a better edge reconstruction capability. Additionally, our model demonstrates adaptive learning capabilities by effectively adjusting to Brownian noise patterns and introduces a sigmoidal noise sequencing method that simplifies training, resulting in faster inference speeds.

## I. Introduction

IN computer vision, image super-resolution (SR) [1] is an important field of study that aims to improve the visual quality of low-resolution (LR) images by increasing their resolution. This study has broad relevance across various fields [2], [3], [4], [5], [6], [7], [8], [9], [10], [11], [12], including surveillance, medical imaging, and the augmentation of images used in daily life. Utilizing the extraordinary ability of deep learning models to recognize complex patterns and capture representations is crucial for extending SR research and application boundaries. This was accomplished by leveraging the exceptional capabilities of deep-learning diffusion models (DM)[13], [14], [15], [16], [17], [18], [19], [20], [21], [22], [23], [24], [25]. This represents a significant step forward in the extraction of high-resolution (HR) features from images.

These algorithms predominantly rely on convolutional neural networks (CNNs) [26] and generative adversarial networks (GANs)[27]. Within the realm of CNN models, SRCNN[28] is considered a groundbreaking approach, followed by VDSR[57], ESPCN[29] and EDSR[30]. These models directly capture the intricate connections between LR and HR images. Conversely, GAN-based models, such as SRGAN[31], BSRGAN[32] and procedure, yielding visually authentic and high-quality super-resolved images. In addition, attention-based[35] models, such as SAN[36] and SRGAT[37] use selective focus processes to efficiently enhance certain portions of an image. Residual networks such as MSRN[38] and regressive networks such as CARN[39] use residual learning techniques to optimize SR processes. Advanced SR algorithms use the transformer[40] and DM[41] as their core architectures. SwinIR[42] exhibits the highest performance among transformer-based models. Subsequently, CFSR[43] and HAT[44] successfully adapted to the SwinIR structure, yielding promising variations and outcomes. The DM-based SR methods have recently demonstrated high generative capabilities. Some methods that have contributed to the field of SR include ResShift[45], DiffBIR[46], CCSR [47], LDM-SR[48], SAM-DiffSR[49], SRDiff[50], StableSR[51], SinSR [88] and InvSR [89]. In our experiment, we compared GAN-based, transformer, and DM-based SR models and found that methods such as CCSR, ESRGAN, and SAM-DiffBIR changed the original outline of the SR image while introducing additional textures that were not present in the ground truth(GT) image. SwinIR and ResShift have a blur effect, whereas DiffBIR and BSRGAN over-smooth the images. Figure 1 shows the visualization results of the SR images generated using state-of-the-art (SOTA) methods compared with GT images. Figure 1 shows that our model not only maintains the original layout but also provides a better resolution.

Thus far, our investigation to resolve these issues extends to the incorporation of stochasticity in Brownian motion (BM) [52], [53] within the DM. The adoption of Brownian noise in SR contributes significantly to the field by amplifying the adaptability [54], [55] and efficacy of the DM-based SR approach. Incorporating Brownian noise allows our model to adaptively learn and adjust its behavior, enhancing image realism. Furthermore, the introduction of a sigmoidal noise sequencing method streamlines the training process, leading to faster inference times. Our contributions to image SR using BM can be summarized as follows:

- We first introduce a novel BM into a DM-based SR, which results in enhanced and more realistic images. The results demonstrate that incorporating the BM significantly improves the realism of the images generated by the SR process.

This work was supported by the National Research Foundation of Korea (NRF) grant funded by the Korea government (MSIT) (No. RS-2023-00217471).

Aradhana Mishra is with CureBay as Chief Data Scientist, Bhubaneswar, India (e-mail: aradhana.mishra@curebay.com).

Bumshik Lee is with the Institute of Energy AI, Korea Institute of Energy Technology, Naju 58330, Republic of Korea (e-mail: bslee@kentech.ac.kr).

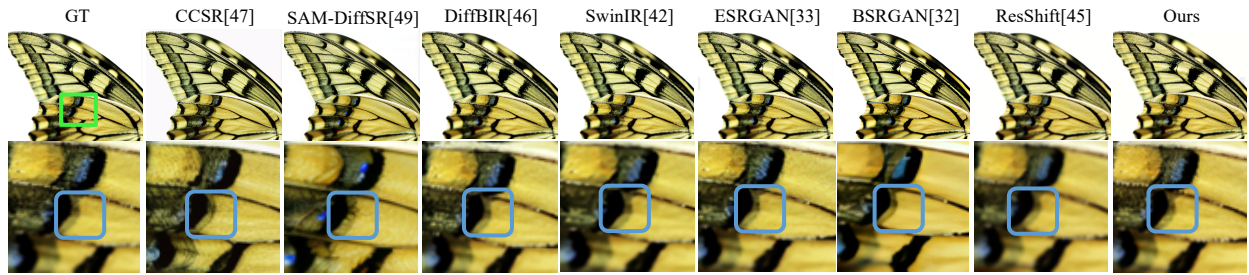

Figure 1. Visualization of SR images generated by SOTA methods. Our method focuses on retaining the original feature as in the GT image with high-quality enhancement. Zoom in and focus on the blue box for clarity.

- A sigmoidal noise sequencing method is introduced to simplify the model training regime, leading to faster inference speed.
- The analysis of the noise patterns in the data demonstrates the adaptive learning capabilities of the proposed model.

The remainder of this paper is organized as follows. Related works are described in Section 2, which provides an overview of existing methods in SR. Section 3 introduces our proposed image SR method, emphasizing the utilization of BM to enhance the learning dynamics of the DM. In Section 4, the experimental results are provided in four subsections: Part A contains the model analysis, Part B presents the evaluation results, demonstrating the effectiveness of our model across various datasets and types of degradation, Part C includes the ablation study, and Part D shows the deoldifying results on old historical images. Finally, in Section 5, we summarize our findings, address the limitations, and explore avenues for future research.

## II. RELATED WORK

SR has evolved from early CNN-based methods, such as SRCNN [28], which enabled hierarchical feature extraction without handcrafted features, to more advanced techniques like VDSR[57] and ESPCN [29]. These earlier approaches are considered SOTA at their time because they established foundational improvements in image reconstruction and introduced end-to-end learning, despite often requiring large datasets and incurring high computational costs.

Transformer-based models, including SwinIR[42] and HAT [44], have advanced the field by leveraging global contextual modeling and long-range dependencies. Their ability to capture both global structures and detailed features has resulted in superior performance over traditional CNNs, which is why these models are regarded as SOTA. However, despite their strengths, Transformer-based approaches can sometimes struggle with capturing fine-grained local details and adapting robustly to diverse degradation conditions, limiting their effectiveness in real-world applications.

To address these shortcomings, DM-based SR techniques have emerged as a promising alternative. Early diffusion approaches, such as ResShift[45] and DiffBIR[46], are considered SOTA because they achieve remarkable image restoration by systematically refining images through iterative denoising processes[56]. These methods effectively mitigate noise, blur, and compression artifacts, thereby producing high-quality, visually realistic outputs. However, their reliance on extensive sampling steps (typically 500–1000 iterations) results in high computational costs, making them impractical for real-time use.

Further innovations, including StableSR[51] and SAM-DiffSR[49], optimize the sampling process, balancing realism with computational efficiency. Techniques like SinSR [88] and InvSR [89] have significantly improved performance by enabling single-step inference, thus making them some of the fastest DM-based methods available. Their SOTA status is attributed to these efficiency improvements and their ability to maintain high-quality reconstructions. Nonetheless, these fast DM-based methods sometimes compromise adaptability due to fixed noise modeling strategies.

To bridge this gap, we introduce PixelBoost, an efficient noise modeling approach that achieves high-fidelity SR in just 10–15 steps. PixelBoost leverages residual noise for rapid convergence and integrates block matching to enhance diffusion stability, mitigating randomness and improving image precision. Unlike traditional methods that rely on static noise priors, PixelBoost dynamically refines noise representation, thus offering superior adaptability, learning efficiency, and computational feasibility across diverse degradation conditions. As demonstrated by LPIPS-based evaluations, PixelBoost outperforms existing diffusion model-based techniques in both visual quality and computational efficiency, thus solidifying its position as a robust, SOTA solution in the SR landscape.

## III. PROPOSED ARCHITECTURE

To address the above-mentioned SR challenges, we propose a DM-based SR model called PixelBoost. The conventional DM utilizes Gaussian noise (GN) modeling for both forward and reverse processes. In contrast, we model the residual [41] of diffusion, which describes the dispersion of particles in a medium by random motion over time. The BM technique [59], [60] enhances this process by increasing the likelihood of particle collisions, allowing substances to mix more thoroughly, and facilitating the transfer of molecules from regions of high concentration to regions of low concentration. Building on this physical principle, we incorporate the BM technique to improve feature representation in deep-learning DM for SR. By integrating information across neural network layers, this method the overall performance of the model in SR. Figure 2 shows the proposed architecture for one full cycle of single-image SR. Let $x_0$ be the HR and $y_0$ the LR images respectively. The residual between the two images is denoted as $\delta_0 = y_0 - x_0$. The BM model can be obtained as $B_t = \alpha_t(\delta_0 \oplus \sigma dW_t)$, where $\oplus$ is an addition operator and $\alpha_t = \eta_t - \eta_{t-1}$ is a drift term. The variable $\eta$ is a component in the shifting sequence $\{\eta_t\}_{t=1}^T$ at time $t$, and $T$ is the final timestep [45]. The shifting sequence with values between 0 to 1 control how much residual $\delta_0$ is included in each timestep in the forward diffusion process. $\sigma dW_t$ is the BM term, where $\sigma$ controls the strength and $dW_t$ is Brownian process[60] for $t$ diffusion step. In the forward diffusion process, a portion of Brownian noise $B_t$ is incorporated at each time step $t$. In the reverse diffusion process, the model gradually reconstructs the details held back by the $B_t$ which contain information about the image, such as edges and fine textures. As shown in Figure 2, the goal is to find the model parameter $\theta$ that minimize $p_\theta(x_{t-1}|x_t, y_0)$ for the reverse transitions as close as possible to the true distribution $q(x_t|x_{t-1}, y_0)$. The model provides a prediction on how to transform a distorted image $x_t$ to clear image $x_{t-1}$. The reverse process follows a U-Net architecture[61] with SWIN transformers (ST) [62], denoted ST U-Net in this paper, sandwiched between the residual blocks in the encoder, decoder, and bottleneck. This ST U-Net architecture is used to translate a noisy image $x_t$ into the next denoised image $x_{t-1}$, continuing this process until a high-quality image $x_0$ is input generated. The proposed architecture employs a sequence of blocks, each specifically tailored to the various phases of feature extraction and processing, as shown in Figure 2. The blocks include convolutional layers for extracting spatial features, residual blocks for learning residuals, and SWIN transformer blocks for capturing intricate patterns using self-attention methods[63] within defined window widths. A detailed description of the proposed architecture is presented in the following section.

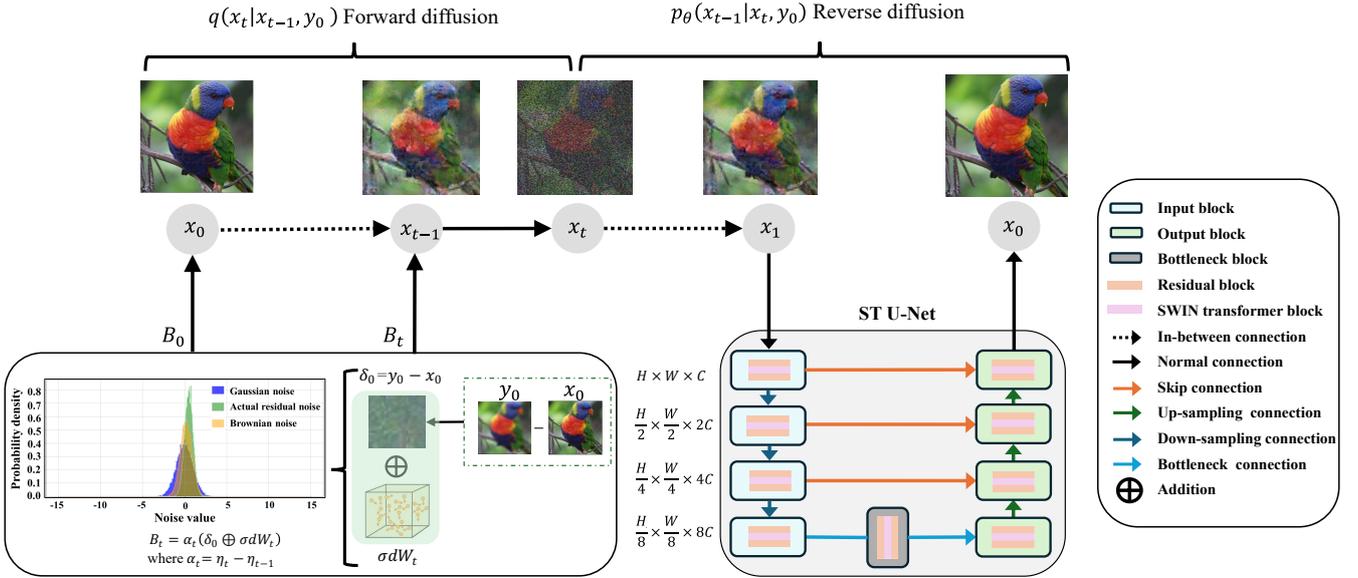

Figure 2. Block diagram of the proposed architecture. It shows a full cycle of DM-based SR for t time steps, which consists of Forward and Reverse processes. The reverse process uses an ST U-Net. Equations (5) and (8) can be referred to for the forward and backward diffusion process and (9) for the Brownian noise sequencing interpretation.

A. Brownian strength

In physics, BM[41], [53] refers to the stochastic displacement of particles in a medium caused by interactions with high-speed atoms or molecules. Unlike conventional DM-based approaches[59] that primarily focus on GN modeling, our method pioneers the explicit integration of BM into the SR process, a concept previously unexplored in the field. This innovative incorporation enables a unique interplay between deterministic and stochastic components, offering a new approach to managing residual noise and image textures effectively. The stochastic nature of BM introduces controlled randomness, promoting diversity and preventing over-smoothing while maintaining perceptual quality. In the proposed PixelBoost model, BM is not just an enhancement but introduces a novel perspective, representing an innovative concept that has been underexplored in super-resolution literature. By explicitly modeling BM in the latent diffusion process, our method bridges the gap between traditional GN approaches and the realistic variability observed in high-resolution images. This results in sharper edges, richer textures, and improved convergence during optimization, avoiding local minima more effectively than standard techniques. Furthermore, the integration of BM within the latent diffusion model (LDM) [22] and residual shifting [19] introduces a stochastic component that is fundamentally different from the deterministic strategies employed by other methods. BM at time $t$ is represented as $BM_t(1)$, following a Gaussian distribution[60] $\mathcal{N}(0, \sigma\Delta t)$, making it a key factor in achieving robust, high-fidelity outputs. This novel approach presents a new perspective on how stochasticity can be leveraged for DM-based SR. Therefore, BM at time $t$ can be represented as $BM_t(1)$.

$$BM_t = \mathcal{N}(0, \sigma\Delta t) = \sigma dW_t, \qquad t = 1, 2, \ldots, T \qquad (1)$$

where, $\sigma$ is the volatile hyperparameter controlling the strength of BM, and $dW_t$ is an infinitesimal increment of the Brownian process, also known as the Wiener process at time $t$. The Wiener process $W_t$ is a continuous-time stochastic process with independent and normally distributed increments, drawn from a normal distribution with mean 0 and standard deviation $\sigma\Delta t$, where $\Delta t$ denotes a small change in time.

The forward diffusion process is represented by determining the conditional distribution of the state $x_t$ with the initial conditions of the previous states $x_{t-1}$ and $y_0$, as shown in Figure 2. The conditional distribution can be determined by using the stochastic differential equation (SDE)[64] in (2).

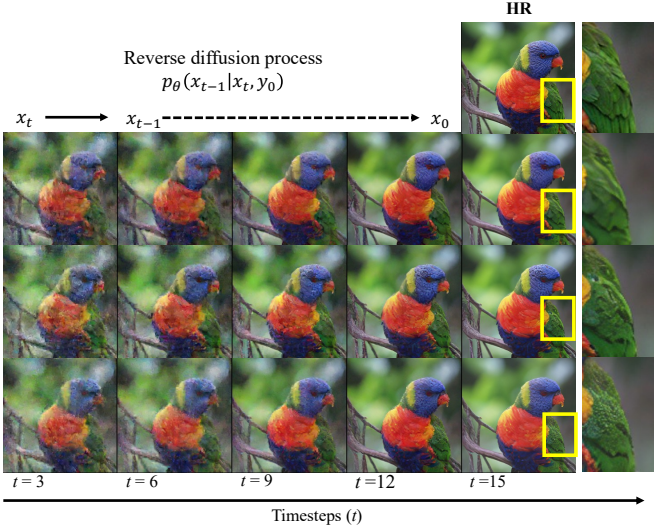

Figure 3. The reverse diffusion process shows noise removal for three different scenarios. First, PixelBoost with σ = 1.5, our proposed model. Second, PixelBoost with σ = 0.01, demonstrating the problem with a low σ value. Third, the base model that is trained on residual without BM. Our proposed model that implements BM can capture better texture and edges, resulting in better pixel value and visuals.

$$dx_t = f(x_t)dt + \sigma dW_t \qquad (2)$$

where $f(x_t)$ is the drift coefficient and is a function of time that influences the average direction in which $x_t$ tends to move, and $\sigma dW_t$ is the increment of BM.

By applying Ito's lemma [64] to it, the mean value of $x_t$ is obtained given the previous condition $x_{t-1}$ as (3).

$$x_t = x_{t-1} + f(x_t) + \sigma W_t \qquad (3)$$

The drift coefficient $f(x_t)$ can be represented in combination with the BM term $B_t = \alpha_t(\delta_0 \oplus \sigma dW_t)$ in (3), as it provides direction for the residual shifting. Therefore, $x_t$ can be expressed as (4). In (1), BM introduces random noise into the diffusion process. This stochastic element, represented by the term $\sigma dW_t$, preserves the inherent unpredictability of the system. Building on this, (4) extends the concept by introducing a directional component to the noise, guided by the function $f(x_t)$, which influences the way the noise evolves over time. While both (1) and (4) include the $\sigma dW_t$ term to maintain the stochastic nature of the model, (4) differentiates itself by incorporating a residual term δ. This δ represents specific adjustments to the baseline model, ensuring that even with the directional influence of $f(x_t)$, the process still follows the same BM framework established in (1). The key difference is that in (1), the noise is purely random, while in (4), it is directed by $f(x_t)$.

$$\begin{aligned} x_t &= x_{t-1} + \alpha_t(\delta_0 \oplus \sigma dW_t) \\ &= x_{t-1} + B_t \end{aligned} \qquad (4)$$

For forward diffusion, the mean of $x_t$ is (4) and $\sigma^2 \alpha_t I$ is the covariance matrix of the Gaussian distribution, where $\sigma^2$ is Brownian noise factor, $\alpha_t$ is the motion drift that increases over time $t$, and $I$ is the identity matrix to ensures that the noise is applied independently in each pixel dimension. Therefore, the final forward equation in the DM can be expressed as a conditional distribution using the BM with means (4) and variance (5).

$$q(x_t|x_{t-1}, y_0) = \mathcal{N}(x_t; x_{t-1} + B_t, \sigma^2 \alpha_t I) \qquad (5)$$
$$t = 1, 2, \dots T$$

where $q$ is the probability distribution of the current state $x_t$ given the information of the previous state $x_{t-1}$ and the initial LR image $y_0$, in the forward diffusion.

The primary concept of the DM is learning the underlying features of an image during the forward process. In the reverse process, the likelihood of obtaining high-resolution images was calculated using the mean and variance. For the reverse diffusion process in SDE, i.e., $dx_{t-1}$, the reverse drift function is taken as $\mu(x_t)$ as (6) [64][41].

$$dx_{t-1} = \mu(x_t)dt + \sigma dW_t \qquad (6)$$

By applying Ito's lemma, the mean value of $x_{t-1}$ can be obtained, given the previous condition $x_t$ in (7).

$$x_{t-1} = x_t + \mu(x_t) + \sigma W_t \qquad (7)$$

From [41], [65], we assume a Gaussian distribution for the reverse diffusion process that can be represented as $p_\theta(x_{t-1}|x_t, y_0)$, where $p_\theta$ is the conditional probability of reconstructing the image from step $x_t$ to $x_{t-1}$ with initial condition LR image $y_0$, parameterized by a neural network parameter $\theta$. The mean of the Gaussian distribution parameterized by $\theta$ can be $\mu_\theta(x_t, y_0, t)$ representing (7)($x_t + \mu(x_t) + \sigma W_t$). The parameter $\theta$ is responsible for injecting Brownian characteristics into the noisy image while denoising. The data-dependent variance of the Gaussian distribution can be represented by $\Sigma\theta(x_t, y_0, t)$. In our model, the neural network used in the reverse diffusion process was ST U-Net. The likelihood of the reverse process can be represented using Brownian mean $\mu_\theta(x_t, y_0, t)$ with variance $\Sigma\theta(x_t, y_0, t)$ as (8).

$$\begin{aligned} p_\theta(x_{t-1}|x_t, y_0) \\ = \mathcal{N}(x_{t-1}; \mu_\theta(x_t, y_0, t), \Sigma\theta(x_t, y_0, t)) \end{aligned} \qquad (8)$$
$$t = 1, 2, \dots, T$$

where $p_\theta$ is the conditional probability distribution for $t = 1, 2, \dots T$. We use the Kullback-Leibler (KL) divergence[13] distributions, $q(x_{t-1}|x_t, y_0)$ and $p_\theta(x_{t-1}|x_t, y_0)$ and minimize the loss using $\min_\theta \sum_t D_{KL}[q(x_{t-1}|x_t, y_0) \| p_\theta(x_{t-1}|x_t, y_0)]$ to make the model to align its predicted distribution $p_\theta$ with the real denoising distribution $q$. During our experiment, we that of both the real residual and the GN models, indicating better performance of our proposed PixelBoost. Figure 3 illustrates the reverse diffusion noise removal process and the addition of the extracted HR features for the three different scenarios. In the first scenario, we analyzed the base model without a BM, which failed to capture detailed patterns and edge definitions. In the second and third scenario, we introduced a BM of strength 0.01 and 1.5, respectively for noise modeling and observed that it shows undesirable artifacts, although it attempts to capture edge definition.

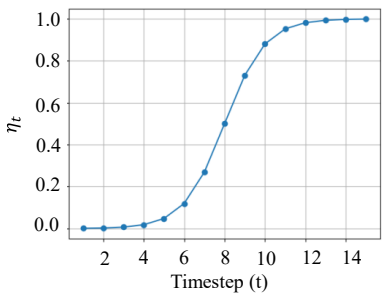

Figure 4: Noise sequence of PixelBoost using timestep-dependent sigmoid function in 15 timesteps.

For illustrative purposes, we present a concise example in one dimension. Let $x_0$ be the initial signal undergoing a forward diffusion process driven by BM. At each timestep $t$ a Gaussian noise increment $\sigma dW_t$ (where $dW_t \sim \mathcal{N}(0,1)$) is scaled by $\alpha_t$ and added to $x$. For instance, if $\alpha_1 = 0.2$, $\sigma = 0.5$, and $dW_1 = -1.2$, then $x_1 = x_0 + 0.2 \times (0.5 \times -1.2)$. Iteratively applying this procedure results in a trajectory that balances stochastic noise with controlled diffusion. In the context of HR image generation, proper BM modeling proves crucial for preserving fine textures and sharp edges. This example clarifies how noise is systematically introduced during the forward diffusion process. Specifically, our empirical findings demonstrate that setting $\sigma \in (0.1, 2)$ within the proposed PixelBoost framework facilitates both robust texture retention and improved edge delineation. By training the model to adapt to time-dependent Brownian noise and minimizing KL divergence $\min_\theta \sum_t D_{KL}$, the neural networks (parameterized by $\theta$) effectively learns to utilize and counterbalance the injected noise. This approach enables it to recognize analogous noise patterns present in real degraded images, thereby enhancing its capacity to restore high-fidelity details while maintaining an appropriate equilibrium between randomness and structure.

### B. Sigmoid noise sequencing

This section describes how Brownian noise is added to an image during forward diffusion. The variable $\eta$ represents a component in the shifting sequence $\{\eta_t\}_{t=1}^T$ as described in [45]. However, it is a complex, non-periodic geometric function. In our proposed model, we formulated $\{\eta_t\}_{t=1}^T$ as a sigmoid function (9) to reduce complexity. During the forward diffusion process in this sequence, where we assume $\eta_1 \to 0$ and $\eta_T \to 1$ calculates the proportion of residual data that is progressively incorporated at each timestep $t$. Figure 4 shows the noise schedule across 15 timesteps ($T = 15$), which illustrates the application of (9) from $t = 1$ to the final step $T$. By modeling with a Brownian residual, the length of the Markov chain is reduced, and the sigmoid sequencing strategy allows us to obtain faster inference time as the complexity of evaluation of the schedule of noise injection is simplified.

BM introduces noise in a statistically stable manner due to its residual properties, enhancing the convergence process during inference. This stability reduces the occurrence of numerical instabilities, minimizing the need for corrective measures that would otherwise increase computation time. As a result, the inference process becomes more efficient and streamlined. Table 2 in the Experimental Results section presents the evaluation results regarding the complexity. The expression of the time schedule is given as $\eta_t$ as shown in (9).

$$\eta_t = \frac{1}{1 + e^{-(t-t_{mid})}}, \qquad t = 1, \dots, T \qquad (9)$$

where $\eta_t$ models a sigmoidal increase from 0 to 1, $t_{mid}$ is the middle of the timestep. The proposed sigmoid noise sequencing improves inference time by offering a more predictable and computationally efficient progression of noise levels compared to the complex, non-periodic sifting sequence described in [45]. This approach facilitates smoother transitions between noise levels, which aids the score network in more accurately estimating gradients. The minor variations in the noise parameter $\eta$ between early and late timesteps are critical for precise score calculation, as they help the model refine its denoising process and preserve fine details. Removing these points could compromise the accuracy of score estimation and result in suboptimal image reconstruction. Thus, maintaining these incremental changes is crucial for enhancing image quality and reducing inference time.

## IV. EXPERIMENTAL RESULTS

**Dataset:** For real-world image SR networks based on a DM, training on large datasets is recommended. Considering this, we chose 1.3 million images from ImageNet[66] as the training dataset. Our experiment involved implementing a ×4 SR to enhance image quality. We trained on an NVIDIA GeForce RTX 3090 for 500,000 iterations. To assess the influence of various forms of noise on our DM, we implemented training techniques that integrate Laplacian, Poisson, and uniform noise other than GN and studied their behavior.

We performed the experiments using traditional testing datasets, such as Set5[67], Set14[68], Urban100[69], B100[70], Manga109[71], and General100[72] which are commonly employed for benchmarking tests of SOTA methods. In addition, we tested historical image datasets to further evaluate the effectiveness of our model and its deoldifying capability. It was also tested on the ImageNet dataset and compared with the BL model. The remainder of this section presents a model analysis that shows the effect of the key components of the model, experimental results with various metric values, visual results, and performance evaluation.

### A. Performance evaluation

In this section, we evaluate the proposed PixelBoost by comparing it with SOTA methods for an SR at a ×4 scale resolution. We compare it with the GAN-based methods, including SRGAN[31], BSRGAN[32] and ESRGAN[33], [34]. Transformer-based methods, including SwinIR[42] and CFSR[43], and DM-based methods, including SAM-DiffSR[49], CCSR[47], DiffBIR[46] and ResShift[45]. All other SOTA methods were re-evaluated under the same experimental conditions in this paper. Experimental conditions in existing SOTA studies vary significantly, which could lead to unfair comparison. To ensure fairness, we performed all experiments under consistent settings, including the dataset, hardware (NVIDIA GeForce RTX 3090), and evaluation metrics used in this paper using skimage.metrics.

Table 1. Quantitative comparison (average PSNR/SSIM/LPIPS/LOE) with SOTA methods for image SR on Set5, Set14, Manga109, B100, Urban100, and General100 testing datasets. The green highlighted ones are the DM-based model.

| Dataset | Set5 | | | | Set14 | | | | Manga109 | | | |
|---|---|---|---|---|---|---|---|---|---|---|---|---|
| Method | PSNR(dB)↑ | SSIM↑ | LPIPS↓ | LOE↓ | PSNR(dB)↑ | SSIM↑ | LPIPS↓ | LOE↓ | PSNR(dB)↑ | SSIM↑ | LPIPS↓ | LOE↓ |
| ESRGAN[33] | 21.68 | 0.8529 | 0.2110 | 268.24 | 23.71 | 0.7856 | 0.2403 | 220.19 | 27.30 | 0.9239 | 0.1172 | 268.28 |
| BSRGAN[32] | 24.69 | 0.8462 | 0.2546 | 191.87 | 24.20 | 0.7829 | 0.3310 | 259.33 | 23.16 | 0.8709 | 0.2241 | 402.36 |
| SRGAN[31] | 21.48 | 0.8409 | 0.2313 | 234.29 | 25.50 | 0.6856 | 0.3245 | 233.46 | 26.18 | 0.8616 | 0.2638 | 322.88 |
| SAM-DiffSR[49] | 24.45 | 0.8888 | 0.2068 | 205.41 | **24.29** | **0.8015** | 0.2563 | 236.13 | 27.64 | 0.9334 | <u>0.1280</u> | 224.36 |
| CCSR[47] | 26.19 | 0.9050 | <u>0.2026</u> | 157.44 | 23.74 | <u>0.7842</u> | 0.2662 | 255.48 | 23.21 | 0.8744 | 0.1963 | 389.66 |
| DiffBIR[46] | 24.36 | 0.8597 | 0.2518 | 182.02 | 24.23 | 0.8010 | 0.3191 | 282.97 | 24.14 | 0.8909 | 0.2105 | 421.20 |
| ResShift[45] | <u>27.07</u> | <u>0.9158</u> | 0.2386 | **146.85** | 23.68 | 0.7113 | 0.3532 | 285.37 | 22.11 | 0.8595 | 0.2466 | 319.13 |
| PixelBoost(Ours) | **27.79** | **0.9216** | **0.1408** | <u>142.48</u> | <u>24.25</u> | 0.7942 | **0.1991** | **217.90** | <u>26.94</u> | <u>0.9236</u> | **0.1190** | <u>300.92</u> |
| SwinIR[42] | 32.92 | 0.9044 | 0.2408 | 251.28 | 29.09 | 0.7950 | 0.3057 | 156.85 | 30.92 | 0.9111 | 0.1716 | 361.98 |
| CFSR[43] | 32.33 | 0.8964 | 0.2476 | 252.68 | 28.73 | 0.7842 | 0.3159 | 161.16 | 30.72 | 0.9151 | 0.1835 | 363.11 |
| Dataset | B100 | | | | Urban100 | | | | General100 | | | |
| Method | PSNR(dB)↑ | SSIM↑ | LPIPS↓ | LOE↓ | PSNR(dB)↑ | SSIM↑ | LPIPS↓ | LOE↓ | PSNR(dB)↑ | SSIM↑ | LPIPS↓ | LOE↓ |
| ESRGAN[33] | 24.65 | 0.8480 | 0.1669 | 363.87 | 22.99 | 0.7550 | 0.1909 | 256.55 | 27.93 | 0.8958 | 0.1568 | 173.25 |
| BSRGAN[32] | 23.54 | 0.8270 | 0.2724 | 343.51 | 21.54 | 0.7191 | 0.2874 | 311.97 | 25.10 | 0.8618 | 0.2582 | 229.75 |
| SRGAN[31] | 24.75 | 0.6400 | 0.2930 | 394.85 | 22.85 | 0.6846 | 0.2233 | 324.89 | 26.13 | 0.8514 | 0.2243 | 234.98 |
| SAM-DiffSR[49] | <u>25.40</u> | 0.8242 | 0.2808 | **289.51** | 22.90 | <u>0.7747</u> | 0.2693 | 266.99 | 22.86 | 0.8528 | 0.2176 | 267.61 |
| CCSR[47] | 24.35 | 0.7973 | 0.2706 | 341.48 | 21.64 | 0.7270 | <u>0.2088</u> | 312.27 | 25.53 | 0.8726 | 0.2169 | 217.15 |
| DiffBIR[46] | 22.48 | 0.8083 | <u>0.2651</u> | 329.24 | 20.71 | 0.6933 | 0.2595 | 340.53 | 24.17 | 0.8514 | 0.2729 | 240.03 |
| ResShift[45] | 24.64 | <u>0.8518</u> | 0.2772 | 302.67 | 22.21 | 0.7407 | 0.2092 | 281.34 | <u>26.63</u> | <u>0.8910</u> | 0.2541 | <u>181.28</u> |
| PixelBoost(Ours) | **25.58** | **0.8573** | **0.1575** | <u>254.95</u> | **23.56** | **0.7443** | **0.1471** | <u>266.23</u> | **27.33** | **0.8954** | **0.1655** | 175.97 |
| SwinIR[42] | 28.13 | 0.9035 | 0.2153 | 281.49 | 25.50 | 0.8353 | 0.1849 | 184.64 | 30.13 | 0.9291 | 0.1859 | 123.77 |
| CFSR[43] | 27.63 | 0.7381 | 0.2649 | 284.03 | 24.68 | 0.8151 | 0.2558 | 204.78 | - | - | - | - |

For quantitative analysis, we measured PSNR[73] and SSIM[74] using the "skimage.metrics" library and LPIPS[75] using "lpips" in the Pytorch environment. LOE[76] measurements assess the naturalness preserved in an SR image by comparing it with the original image. Table 1 presents a comprehensive comparison across multiple benchmark datasets, including Set5, Set14, Manga109, B100, Urban100, and General100, where PixelBoost consistently outperforms SOTA DM-based SR models. Although Transformer-based SOTA methods [42], [43] and [44] achieve higher PSNR values, they often fail to retain perceptual fidelity, whereas PixelBoost demonstrates significantly lower LPIPS and LOE scores, indicating superior preservation of fine textures and more realistic image reconstructions.

The qualitative superiority of PixelBoost is further demonstrated in Figure 5, which illustrates SR results for ×4 resolution on Set14, highlighting its ability to preserve intricate textures and structural details. In contrast, existing SOTA methods introduce excessive smoothing and texture loss, leading to hazy reconstructions. PixelBoost effectively maintains high-frequency details, ensuring sharper and more perceptually accurate outputs. As observed in Figure 5, PixelBoost can preserve the original texture and fine details without sacrificing the high-definition edges when examining the zoomed-in patch. In contrast, SOTA methods often smooth out details excessively, leading to texture loss and hazy results.

Table 2 presents a comparison of the computational complexity between the proposed PixelBoost and the SOTA methods on the Set5 dataset in sampling steps, inference times, and number of parameters. The values in Table 2 represent the optimal parameters for achieving the best performance in each method. Table 2 shows that PixelBoost requires the same number of sampling steps as DM-based models such as CCSR[47] and ResShift[45]. However, as evidenced by Table 1 and Figure 5, our proposed method achieves significantly higher visual quality under the same sampling conditions. This is because, unlike existing DM-based models that require higher computational resources and parameter counts, PixelBoost integrates BM to enhance the diffusion process with only about 10–15 sampling steps, without increasing model complexity.

DM-based SR methods were also tested in the real-world dataset, such as General100 [72] and ImageNet [66]. Figure 6 and Figure 7 show the comparative visual results with LPIPS values for ImageNet and General100 datasets, respectively. As shown in Figure 6, our proposed model PixelBoost shows the lowest LPIPS values for both datasets. In Figure 7, focusing on the General100 dataset, we can see that PixelBoost excels at capturing fine details. For instance, the water droplets on flower petals in the output of our model are remarkably close to the GT. Other DM-based SR methods tend to either blur these intricate textures or introduce noticeable artifacts. In contrast, PixelBoost produces sharper and more true-to-life results. This improvement in visual quality is particularly evident in how well it preserves texture details, which aligns with the lower LPIPS scores observed for PixelBoost.

In summary, the technical contributions of this study are demonstrated through quantitative and qualitative evaluations, establishing the superiority of PixelBoost over existing SOTA methods in SR. PixelBoost significantly improves efficiency, achieving high-quality SR in just 10–15 sampling steps, reducing inference time while maintaining superior image fidelity. The integration of BM enhances adaptability, ensuring more stable and accurate noise refinement without increasing computational complexity.

B. Model analysis

1. Noise pattern analysis

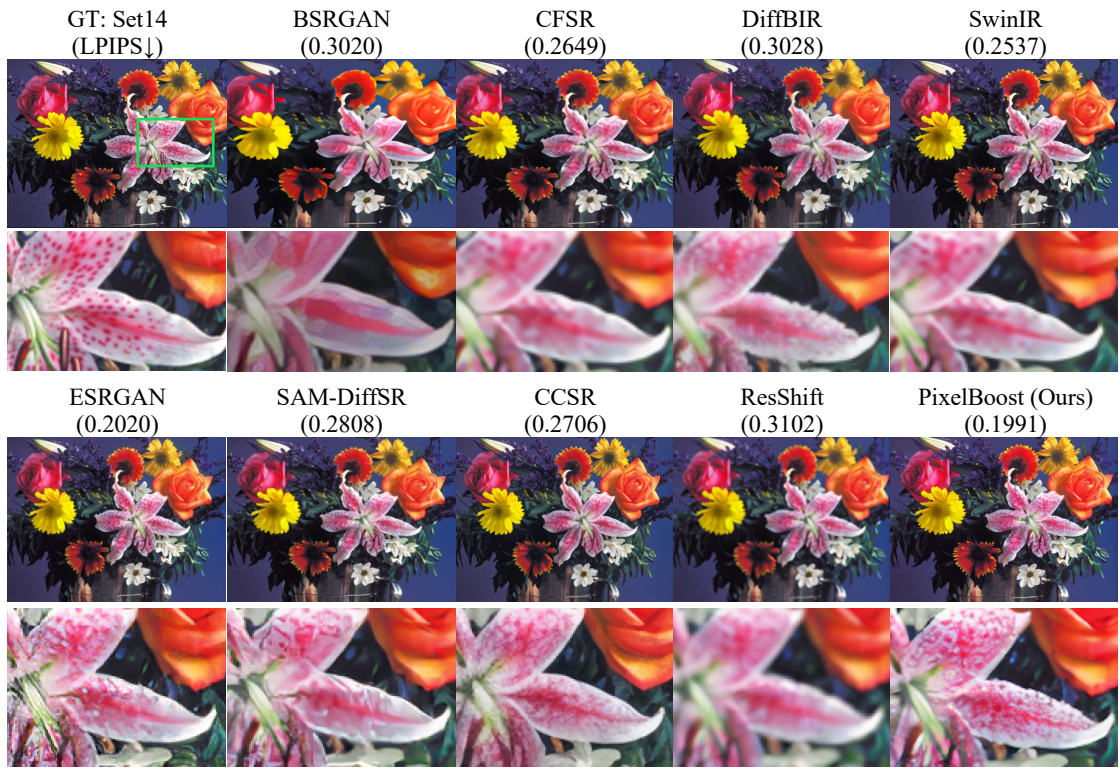

Figure 5. Visual results for ×4 resolution with SOTA methods in Set14 classical dataset. These datasets include images such as animals, human faces, and flowers.

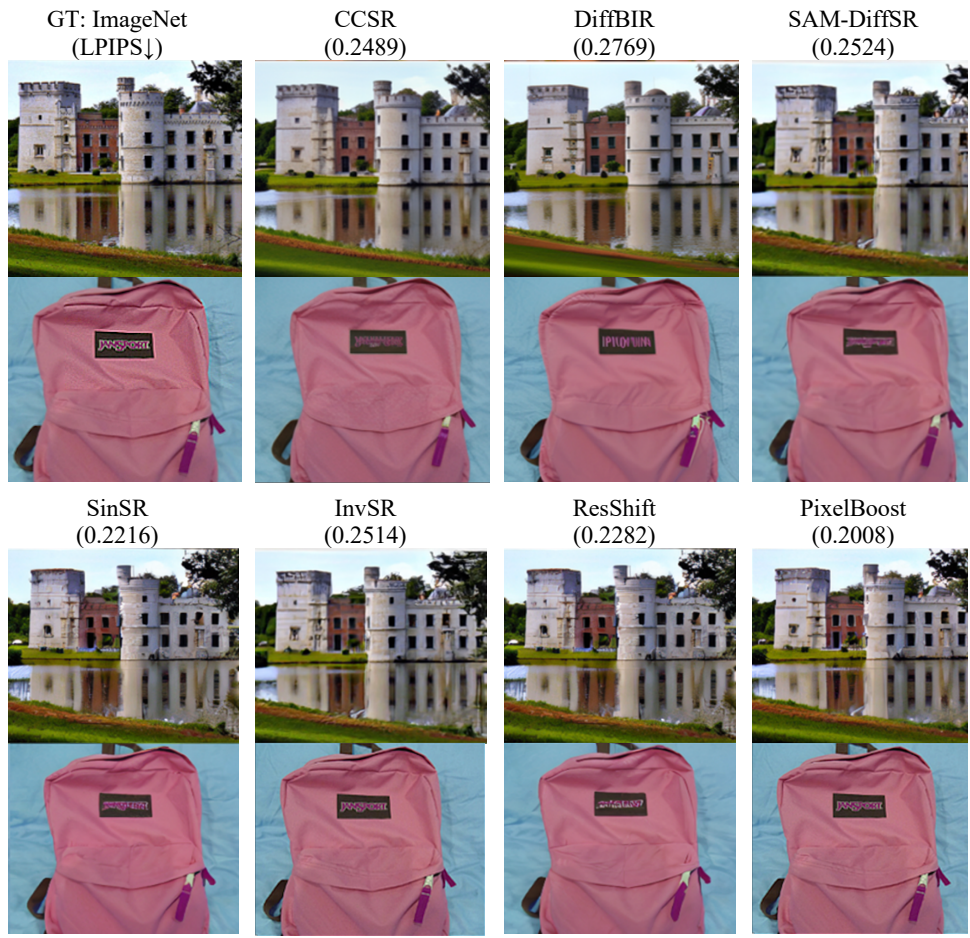

Figure 6. Visual results for ×4 resolution with SOTA methods in ImageNet testing real-world dataset.

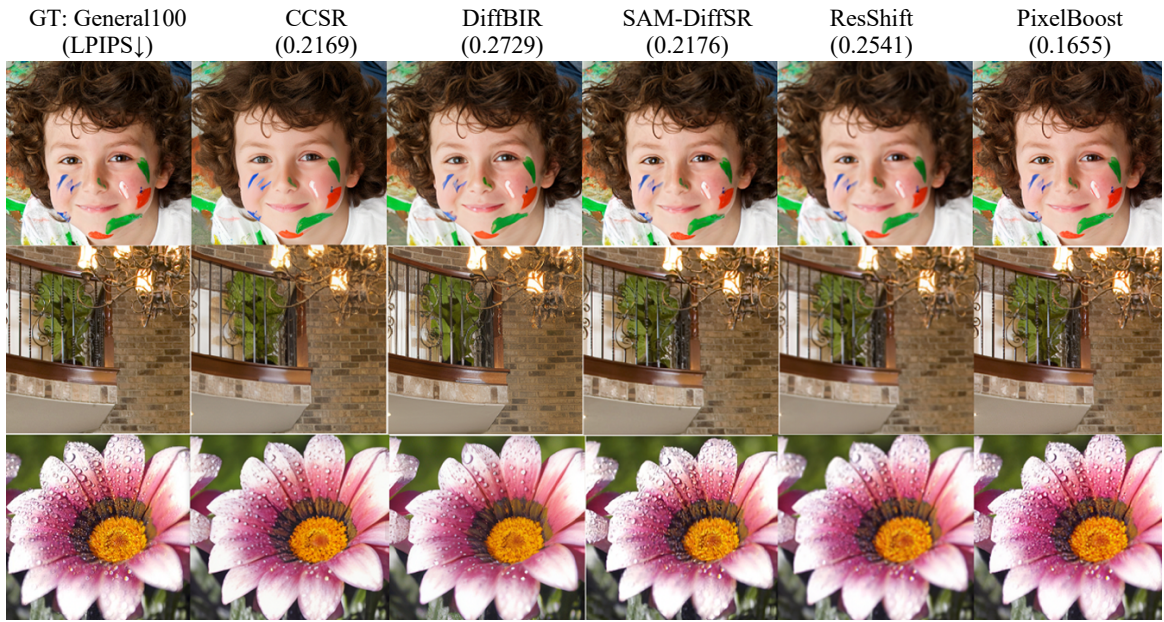

Figure 7. Visual results for ×4 resolution with SOTA methods in General100 dataset

Table 2. Model complexity comparisons of PixelBoost to SOTA on the Set5 dataset

| Methods | LDM-SR[48] | StableSR[51] | CCSR[47] | DiffBIR[46] | PASD[58] | ResShift[45] | PixelBoost |
|---|---|---|---|---|---|---|---|
| Sampling steps | 200 | 200 | 15 | 50 | 20 | 15 | **15** |
| Inference time(s)/ Image | 5.21 | 18.90 | 2.22 | 5.85 | 6.11 | 1.12 | **1.09** |
| # of parameters (M) | 168.95 | 1409.11 | 1700.91 | 1716.71 | 1883.70 | 173.91 | **118.59** |

Although both GN[77] and Laplacian noise (LN)[78] distributions are symmetrical, GN performs better than LN because GN has fewer extreme outliers than LN when incorporating them into the DM. The Poisson noise (PN) is known to be characterized by non-negative integer values and a skewed distribution. In contrast, the Uniform noise (UN) [79] has a constant probability distribution across its defined range. Although GN offers certain advantages over LN, PN and UN, we compared its behavior with the performances of LN, PN and UN. Figure 8(a) shows the variations of PSNR values over the epoch for each noise. The lower PSNR and higher LPIPS values indicate that the model lost its structural integrity, and unnecessary noise is introduced when the model is trained with PN, UN, and LN. In contrast, Gaussian noise shows the highest PSNR and lowest LPIPS values for the DM. It also justifies the theory described in [80].

However, since the Brownian noise is introduced in our proposed model, it is worthwhile to examine how the Brownian noise is fitted to the actual noise. To do this, the chi-square hypothesis test [81] was performed. The chi-square test statistic quantifies the difference between the observed and expected data in categorical data. In this context, the null hypothesis assumes that the expected actual residual noise distribution is the same as the noise distribution generated by observed Brownian noise, GN, LN, and PN. Figure 8(b) shows the probability density of GN and Brownian noise compared with actual residual noise.

Therefore, according to Table 3, the comparison between the actual residual noise and BN yields the lowest chi-square test statistic (0.1541), indicating the closest match between the actual residual noise distribution and the Brownian noise distribution. Conversely, the comparison between the actual residual and PN exhibits the highest chi-square test statistic (0.8688), indicating a relatively lower fit between the two distributions. The lower values of the chi-square test statistic indicate a better fit between the observed and expected frequencies.

Table 3. Chi-Square test statistic comparison for actual residual vs. various noise types.

| Comparisons | Test statistic↓ |
|---|---|
| Actual residual vs Brownian | 0.1541 |
| Actual residual vs Gaussian | 0.1763 |
| Actual residual vs Laplacian | 0.5372 |
| Actual residual vs Poisson | 0.8688 |

2. Brownian analysis

The performances of the proposed model at different Brownian strength ($\sigma$) values are analyzed. When the Brownian strength is significantly low (e.g., 0.01), the model shows granular results that lack details of the texture. This is because the fine details that are important for the image are blurred and sharpness is lost. We also observed that potential artifacts and unnatural patterns were introduced into the HR output images. In contrast, the low $\sigma$ leads to high LPIPS value due to the loss of structural information. Even a minor deviation caused by lower-strength BM led to a disproportionally high LPIPS score.

As BM operates by diffusing pixel values over neighboring regions with low $\sigma$, it does not adequately smooth out the high-frequency components, such as noise and small-scale artifacts. Rather, it randomly distributes high-frequency variations, thereby increasing the prominence of the output image. A Brownian strength of 0.01 shows the lowest performance across all testing datasets, was included in our experiment. For different values, we performed experiments on the standard Set5[67], Set14[68], Urban100[69],

B100[70], Manga109[71], General100[72], Historical[66], Real Old[82] and ImageNet[66] datasets. Table 4 shows the Table 4. Quantitative results of the model trained for different $\sigma$ values to find the optimum range for the best results in terms of realism, texture, and edge sharpness.

| Data set | Set5 | | | ImageNet | | |
|---|---|---|---|---|---|---|
| BM ($\sigma$)↑ | PSNR (dB)↑ | SSIM ↑ | LPIPS ↓ | PSNR (dB)↑ | SSIM ↑ | LPIPS ↓ |
| 0.01 | 22.34 | 0.8019 | 0.3956 | 19.45 | 0.6821 | 0.5672 |
| 0.4 | 27.49 | 0.9187 | 0.1426 | 22.92 | 0.7008 | 0.2912 |
| 0.5 | 27.14 | 0.9077 | 0.1512 | 22.80 | 0.7198 | 0.2878 |
| 0.6 | 27.68 | 0.9179 | 0.1447 | 22.87 | 0.7291 | 0.2776 |
| 0.9 | 27.49 | 0.9179 | 0.1445 | 22.96 | 0.7498 | 0.2355 |
| 1.0 | 27.52 | 0.9158 | 0.1632 | 23.12 | 0.7522 | 0.2113 |
| 1.2 | 27.78 | 0.9188 | 0.1417 | 23.67 | 0.7632 | 0.2091 |
| **1.5** | **27.79** | **0.9216** | **0.1408** | **23.85** | **0.7783** | **0.2008** |
| 5.0 | 27.60 | 0.9183 | 0.1437 | 23.33 | 0.7712 | 0.2024 |
| 10.0 | 26.50 | 0.9068 | 0.1754 | 23.09 | 0.7689 | 0.2374 |

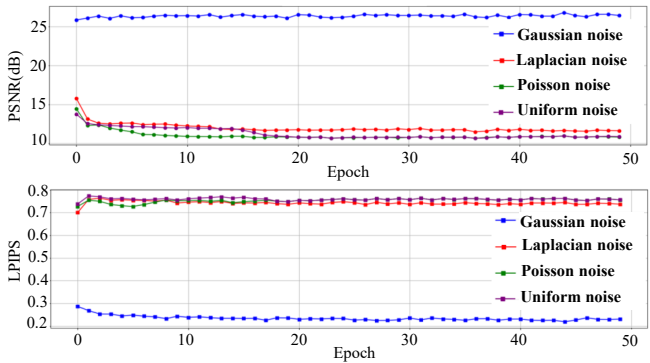

(a)

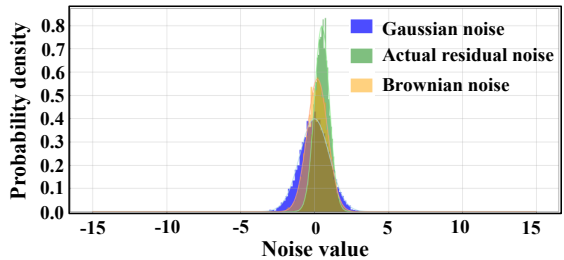

(b)

Figure 8(a). Training progression showing PSNR and LPIPS values for model training using GN, LN, PN, and UN with $\sigma = 1.5$. Testing results in Set5 in the format Noise (PSNR[dB]↑/SSIM↑/LPIPS↓): Gaussian (27.79/0.9216/0.1408), Laplacian (11.04/0.2657/0.7260), Poisson (10.23/0.2322/0.7555) and Uniform (12.30/0.3269/0.7192). (b). Probability density of GN, actual residual noise, and Brownian noise.

quantitative results of the model trained for different $\sigma$ values for dataset set5 and a subset testing dataset taken from ImageNet. It is shown that the most enhanced and realistic images are achieved from the models when the $\sigma$ is between 0.1 and 2, as shown in Table 4 for the Set5 dataset. An additional higher value does not degrade the image quality but causes a slight drop in the metric values.

In Figure 9, we present the training progress and visual outcomes of our models. Figure 9(a) shows the LPIPS and PSNR metrics for three different values of Brownian strength throughout the training period. The data indicates that, while the training nature remains consistent, each model achieves higher metric values, particularly around $\sigma = 1.5$, suggesting that varying Brownian strength influences performance and convergence. Figure 9(b) showcases the visual results of the models, illustrating how different Brownian strengths affect the quality of the super-resolved images, along with the quantitative metric values for the testing dataset Set5, highlighting performance differences among the models.

C. Ablation study

In the ablation study, we performed experiments to investigate the effect of the BM in the DM. In addition to PSNR, SSIM, LPIPS, and LOE, additional metrics such as Multi-Scale Structural Similarity Index for Image Quality Assessment (MUSIQ)[83] and Fréchet Inception Distance (FID)[84] were also evaluated. We evaluated these metrics and compared them with the BL model. The BL model comprises a LDM with ResShift's residual noise concept. Qualitative and quantitative tests were performed in all standard SR testing datasets for ablation studies. Table 5 shows comparative results on the BL (LDM with ResShift's residual concept) with the PixelBoost (BM) for Set5 dataset. As shown in Table 5, the PixelBoost achieves better performance in PSNR, SSIM, MUSIQ, FID, and LOE than BL. The visual results of the ablation study are presented in Figure 10, alongside a pixel value evaluation. These results clearly demonstrate that the proposed PixelBoost achieve significant enhancement. We also performed an edge analysis of the generated SR images between the PixelBoost and BL models. We created 7×7 patches and obtained the gradient

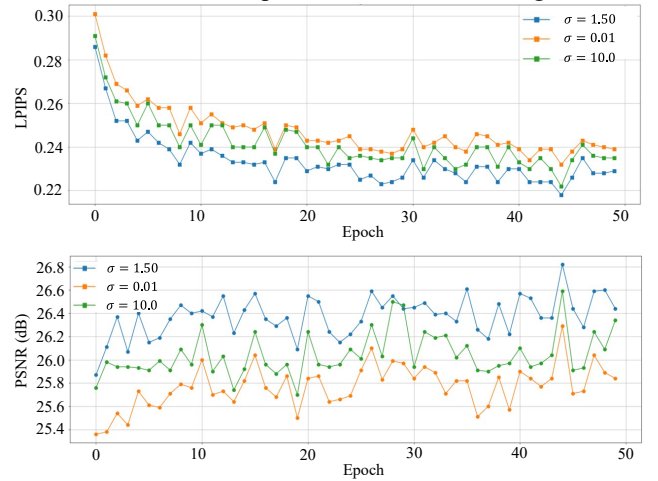

(a)

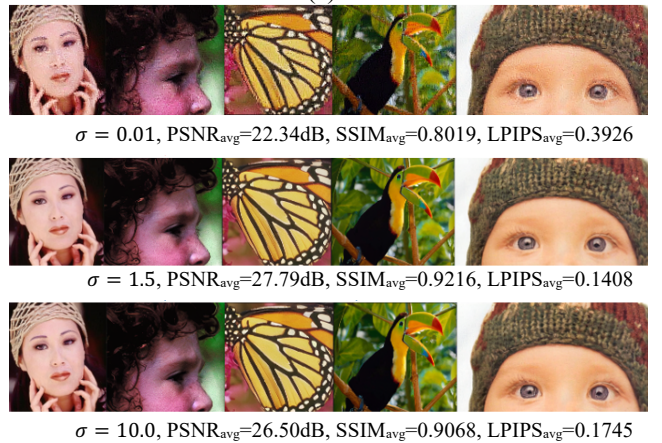

$\sigma = 0.01$, PSNR$_{avg}$=22.34dB, SSIM$_{avg}$=0.8019, LPIPS$_{avg}$=0.3926

$\sigma = 1.5$, PSNR$_{avg}$=27.79dB, SSIM$_{avg}$=0.9216, LPIPS$_{avg}$=0.1408

$\sigma = 10.0$, PSNR$_{avg}$=26.50dB, SSIM$_{avg}$=0.9068, LPIPS$_{avg}$=0.1745

(b)

Figure 9. (a) Training performance in LPIPS and PSNR (dB) for the model trained on $\sigma \in (0.01, 1.5, 10)$ over 500,000 iterations. (b) Visual × 4 SR results of the model trained on $\sigma \in (0.01, 1.5, 10)$ tested on the Set5 dataset.

magnitude values for each patch using the Sobel operation[85]. Figure 11 shows the edge maps for BL and PixelBoost, and their difference values for the gradient magnitude in each patch. Higher gradient magnitudes reflect stronger changes in intensity between neighboring pixels, which correspond to clearer and more detailed edges. Therefore, the PixelBoost model gives the clarity and sharpness of edges in the SR images compared to the BL, leading to a more detailed and visually appealing result.

### D. DeOldify

The proposed PixelBoost achieves robust performance with real-world and synthetic datasets, as shown in the previous chapter. We investigate that the proposed method can also show promising performances for image restoration for old and historical images. "DeOldify" is a term used for the restoration of old historical images and videos. This capability marks a significant achievement because historical datasets typically present unique challenges owing to their varying qualities and the presence of artifacts from older photographic techniques. PixelBoost's ability to enhance such images and increase their resolution while maintaining and recovering the original details shows its versatility and effectiveness across various data types as shown in Figure 12. This performance underscores the potential of our model as a valuable tool for digital restoration and preservation efforts, where enhancing the clarity of historical images can provide better insight and accessibility to past visual records. However, it still lacks the capability to restore heavily corrupted data, as demonstrated by the networks in references [82], [85], [86] and [87].

## V. CONCLUSION

In this study, we propose a low-complexity and high-performance image SR model that leverages the DM framework, specifically integrating BM to inject stochasticity for a better diffusion process. This approach enhances the unpredictability and randomness of the generation process, making the Brownian noise distribution closely align with the HR image distribution. Consequently, our method produces high-quality, realistic, super-resolved images through adaptive learning. We also introduce a simplified noise schedule using a sigmoidal approach, which reduces inference time without compromising image quality. Overall, we achieved higher quality SR images as compared to SOTA methods while making the DM less computationally expensive. For future research, dynamic Brownian strength estimation during model training would be beneficial and expanding the training dataset can help make the model more general.

Table 5. Ablation studies quantitative results performed for Set5 dataset.

| Methods | #parameters (M) | Inference time (s) | PSNR (dB)↑ | SSIM↑ | LPIPS↓ | MUSIQ↑ | FID↓ | LOE↓ |
|---|---|---|---|---|---|---|---|---|
| **BL** | 173.91 | 1.12 | 27.07 | 0.9158 | 0.2386 | 54.62 | 185.43 | 146.85 |
| **BL+ BM** | 118.59 | 1.09 | 27.79 | 0.9216 | 0.1408 | 72.16 | 73.32 | 142.48 |
| **Δ** | **-55.32** | **-0.03** | **+0.72** | **+0.0058** | **-0.0978** | **+17.54** | **-112.11** | **-4.37** |

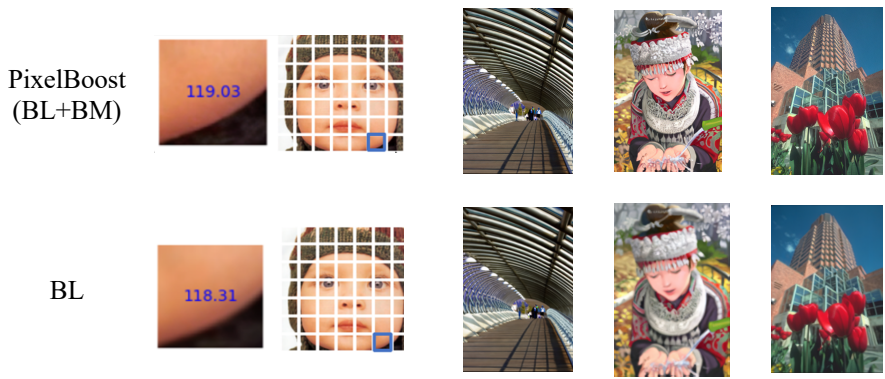

Figure 10. Pixel value evaluated and compared for SR-generated images between BL and PixelBoost (BL+BM)

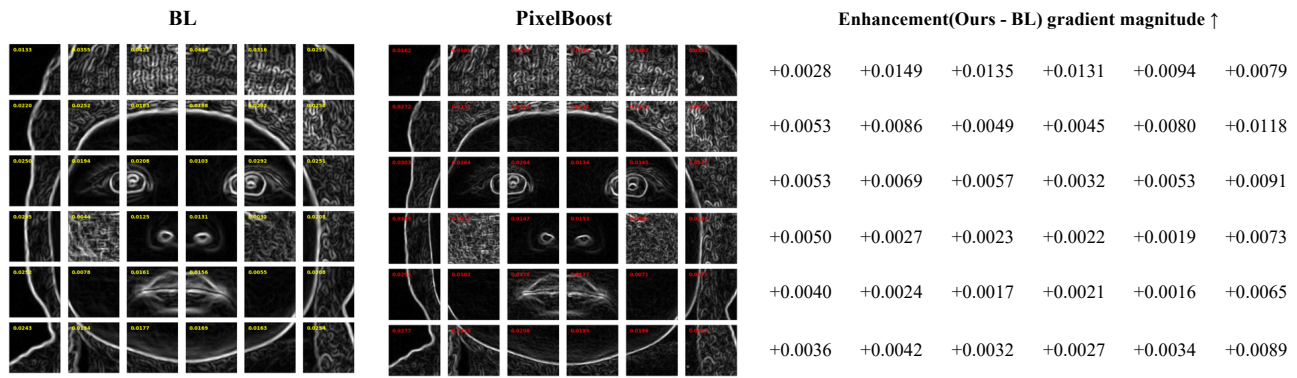

Figure 11. Edge analysis results between PixelBoost and BL in 7×7 patches. The gradient magnitude is increased.

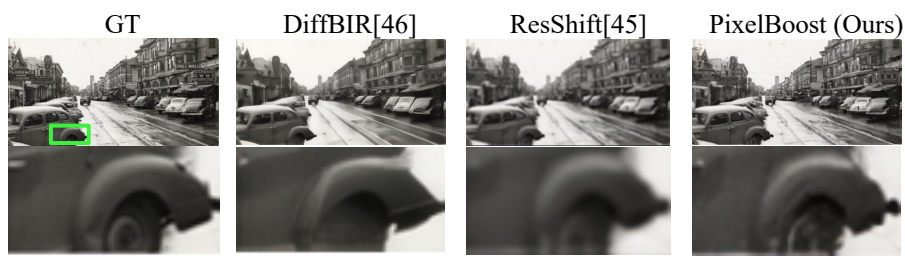

Figure 12. Visual results on the historical dataset for the proposed PixelBoost and two other existing DM.

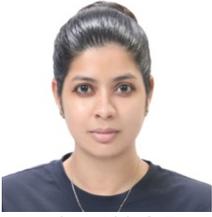

**Aradhana Mishra** received her B.Tech. degree from KIIT University, Bhubaneswar, India, in 2009. She gained industry experience at Aricent, TCS, and Truminds before completing her M.S. in Information and Communications Engineering at Chosun University, South Korea, 2022-2024. Since 2024, she has been the Chief Data Scientist at CureBay, leading AI-driven healthcare projects in contactless vitals monitoring, mental health, dental diagnostics, and women's wellness. Her research focuses on deep learning, generative AI, and audio, video, and image processing for real-time clinical applications.

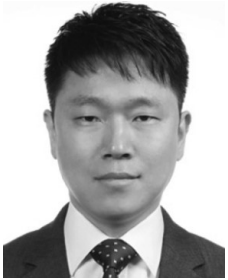

**Bumshik Lee** (M'07) received the B.S. degree in Electrical Engineering from Korea University, Seoul, South Korea, in 2000, and the M.S. and Ph.D. degrees in Information and Communications Engineering from Korea Advanced Institute of Science and Technology (KAIST), Daejeon, South Korea, in 2006 and 2012, respectively. He was a Research Professor at KAIST, in 2014, and a Post-Doctoral Scholar at University of California, San Diego (UCSD), CA, USA, from 2012 to 2013. He was a Principal Engineer at the Advanced Standard R&D Laboratory, LG Electronics, Seoul, from 2015 to 2016. From September 2016 to June 2025, he was with the Department of AI Software, Chosun University, Gwangju, South Korea as a Faculty member. Since July 2025, he has been with the Institute of Energy AI, Korea Institute of Energy Technology (KENTECH), as a faculty member. His research interests include video processing, video security, and medical image processing.